\documentclass[letterpaper,10 pt, conference]{ieeeconf}  

\IEEEoverridecommandlockouts                              

\usepackage{graphicx} 
\usepackage{amsmath} 

\title{\LARGE \bf
End-to-End 3D-PointCloud Semantic Segmentation for Autonomous Driving
}

\author{
 \textbf{Mohammed Abdou}$^*$\\
 Valeo, Egypt\\
 \text{Mohammed.Abdou@valeo.com}\\\\
 \and
 \textbf{Mahmoud Elkhateeb}$^*$\\
 Valeo, Egypt\\
 \text{Mahmoud.Elkhateeb@valeo.com}\\\\
 \and
 \textbf{Ibrahim Sobh}\\
 Valeo, Egypt\\
 \text{Ibrahim.Sobh@valeo.com}\\\\
 \and
 \hspace{210pt}\textbf{Ahmad El-Sallab}\\
 \hspace{210pt}Valeo, Egypt\\
 \hspace{210pt}\text{Ahmad.El-Sallab@valeo.com}
\thanks{$^*$ indicates equal Contributions}}

\begin{document}

\maketitle

\begin{abstract}
3D semantic scene labeling is a fundamental task for Autonomous Driving. Recent work shows the capability of Deep Neural Networks in labeling 3D point sets provided by sensors like LiDAR, and Radar. Imbalanced distribution of classes in the dataset is one of the challenges that face 3D semantic scene labeling task. This leads to misclassifying for the non-dominant classes which suffer from two main problems: a) rare appearance in the dataset, and b) few sensor points reflected from one object of these classes. This paper proposes a Weighted Self-Incremental Transfer Learning as a generalized methodology that solves the imbalanced training dataset problems. It re-weights the components of the loss function computed from individual classes based on their frequencies in the training dataset, and applies Self-Incremental Transfer Learning by running the Neural Network model on non-dominant classes first, then dominant classes one-by-one are added. The experimental results introduce a new  3D point cloud semantic segmentation benchmark for KITTI dataset.
\end{abstract}

\section{INTRODUCTION}

One of the core challenges with intelligent transportation systems is realizing an accurate perception of the surrounding environment that is essential for safe autonomous driving. Recently, perception through scene labeling is considered as a fundamental task for robotics \cite{tchapmi2017segcloud} \cite{ku2017joint} \cite{chen2017multi} \cite{qi2017frustum}. Most approaches depend on camera sensor \cite{badrinarayanan2017segnet} \cite{barkau1996unet} to output segmented images for the detected objects, while others depend on different sensors like LiDAR, and Radar \cite{zhou2017voxelnet} for automotive applications. These sensors produce a sparse 3D point cloud which can be defined as a set of points that describes a 3D space. In autonomous driving, 3D scene labeling is most commonly used in classifying and detecting objects in the surrounding environment, like cars, trucks, pedestrians, trees, buildings, cyclists, traffic signs, and so on.

Imbalanced datasets problem is one of the most difficult challenges that faces not only semantic segmentation tasks but also machine learning applications in general. Imbalanced datasets, are biased towards dominant classes rather than others from the perspective of having higher number of instances. Likewise, imbalanced point cloud datasets also contain fine-grained classes which have fewer reflected points from the class object than coarse-grained classes. These reflected points are generated in case of using point-sensors like LiDARs, and Radars. As an example from automotive field, suppose using LiDAR sensor, number of instances for car and truck classes are greater than others, in addition their number of points reflected are much greater than others. The two previously mentioned problems may lead to either miss or wrongly classifying whole classes points.

\begin{figure*}
\centering
\fbox{\includegraphics[width=1\linewidth]{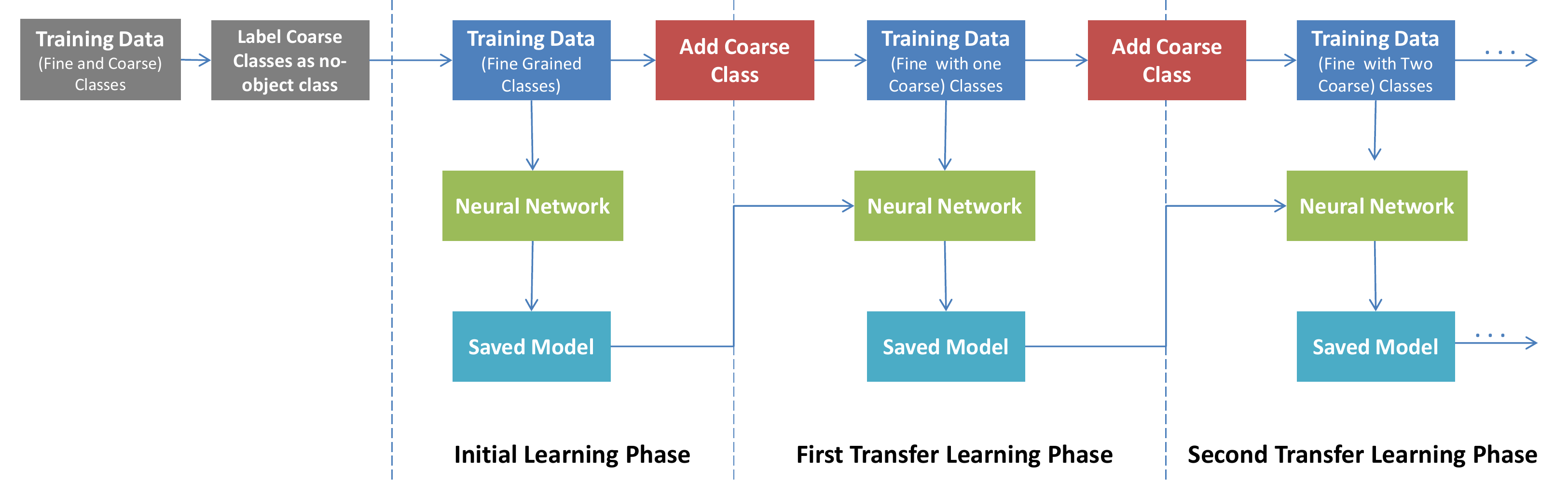}}
\centering
\caption{Incremental Learning Technique}
\label{fig:Incremental_Learning}
\end{figure*}

Our proposed methodology \textit{Weighted Self-Incremental Transfer Learning} solves the imbalanced point cloud datasets problems. It depends on merging two main methods: a) weighted loss function, and b) self incremental transfer learning. Weighted loss function is used to give high weights for the non-dominant fine-grained classes rather than dominant coarse-grained classes as a penalization factor. Classes weights are calculated based on the two main problems of the imbalanced point cloud datasets: a) number of appearance instances for each class, and b) number of points reflected from each class in the dataset. Additionally, Self incremental transfer learning depends on applying training-phase firstly on non-dominant fine-grained classes until having a sufficient representative model. After that it incrementally adds one-by-one dominant coarse-grained class, then applying training-phase depending on the previous pre-trained model until having a sufficient model for the whole imbalanced point cloud dataset. Self-Incremental transfer does not depend on any pre-trained model, but our model is learned continuously as shown in Figure \ref{fig:Incremental_Learning}. 

It is experimentally demonstrated that using weighted loss only, or self-incremental transfer learning only do not solve the problems of imbalanced point cloud datasets like KITTI dataset \cite{Geiger2013IJRR}. PointNet++ \cite{haopointnet++}, is the baseline of our experimental results, applied on KITTI dataset. In consequence, it misclassifies KITTI's non-dominant fine-grained classes: pedestrians, and cyclists. However, our methodology not only classifies them correctly, but also enhances mean IoU measurement metric of point-wise classification compared with applying each technique alone (i.e. weighted loss only or self-incremental transfer learning only), as shown in the experimental results section.

\section{Related Work}


Learning from imbalanced datasets \cite{he2008learning} is an important topic, arising very often in practice in classification problems, that may lead to misclassify most of the data as the dominant class. Many approaches \cite{sun2009classification} \cite{haixiang2017learning} are developed across different levels solving learning from imbalanced datasets. Data-level approaches \cite{sun2009classification} depend either on under-sampling technique for dominant classes or over-sampling technique for the non-dominant classes. Random Under-sampling \cite{anand2010approach}, and Over-sampling techniques \cite{kotsiantis2003mixture} \cite{cao2013integrated} are non-heuristic method that aims to balance class distribution through the random elimination of majority class examples, and through the random replication of minority class examples \cite{chawla2002smote} by generating new synthetic data respectively. Cost sensitive Learning level approach \cite{sun2009classification} \cite{haixiang2017learning} encodes the penalty of misclassifying in terms of weights like weighted loss function. Ensemble approaches \cite{sun2009classification} such as Boosting method \cite{galar2012review} aims to construct multiple models from the original data and then aggregate their predictions when classifying unknown samples. Focal Loss \cite{lin2018focal} applies a modulating term to the cross entropy loss in order to focus learning on non-dominant fine-grained classes.

Incremental Learning is a dynamic technique either for supervised and unsupervised learning which aims to extend the model's knowledge. This technique is applied when training data is available gradually over time or its size is out of system memory limits. This means that the model is fed with different small sized data incrementally through time which may lead to forgetting the firstly learned data. Initial methods \cite{castro2018end} targeted the SVM classifier which encodes the classifier learned on old data to learn the new decision boundary together with new data added. Incremental machine learning algorithms are modified to adapt with the new added data solving forgetting problems \cite{goodfellow2013empirical}. However, it differs from our proposed technique as our model is fed with the same training data each step with gradual increase in the supported classes. 

Recent approaches convert the 3D point cloud into different representation before feeding it into a Deep Neural Network \cite{zhou2017voxelnet} \cite{maturana2015voxnet}. This is because 3D point cloud points are not in a structured format, so they transformed the data into 3D regular voxel-grids \cite{zhou2017voxelnet} \cite{maturana2015voxnet}. This representation results in having unnecessarily volumes of data which lead to heavy computations. Other approaches use 3D point cloud points directly as an input to the Deep Neural Network \cite{qi2017pointnet} \cite{haopointnet++}. PointNet \cite{qi2017pointnet} is a unified architecture that takes point cloud points directly with the ground truth point wise labels as inputs. However, it fails in capturing and extracting local structure features for complex sparse 3D points in the scene. As an extension for PointNet architecture to solve its issues, PointNet++ \cite{haopointnet++} is introduced to have the ability of extracting deep local features. However, based on our experimental results, PointNet++ also suffers from misclassifying imbalanced point cloud datasets.


The following section introduces the paper methodology which describes the main contributions in details. After that, section \ref{Exper_Res} describes the imbalanced point cloud dataset used, explains experimental setup, and the main measurement metric used. Experimental results for applying our methodology are described in section \ref{Exper_Results}. Main conclusions are described in section \ref{conclusion}.

\section{Methodology}
\label{methodology}
This section describes Weighted Self-Incremental Transfer Learning proposed methodology for solving the imbalanced point cloud dataset problems.

\subsection{Weighted Self-Incremental Transfer Learning}
\label{Section_label}
Our proposed solution is to merge and apply both techniques: weighted Loss Function with self-incremental transfer learning together obtaining a \textit{new generalized methodology} capable of learning from imbalanced training datasets contains non-dominant fine-grained classes. Self-Incremental learning procedure is followed in addition to weighted loss function weighting penalization technique for the wrongly classified points.

\subsubsection{Weighted Loss Function}
\label{Weighted_Loss_Function}
Cross entropy loss function penalize the misclassification for the whole classes equally. 
\begin{equation}
Loss = - \sum^{N}_{i} {L_i} \log({S_i})
\end{equation}
where ${N}$ is the number of classes, ${S_i}$ is the output predicted probabilities from softmax layer, and ${L_i}$ is one hot encoded ground truth labels. This means that one point misclassified from non-dominant class is penalized exactly as one point from dominant class which is unfair. Weighted cross entropy loss \cite{badrinarayanan2017segnet} is most commonly used in order to tackle imbalanced dataset problems with non-dominant classes.

\begin{equation}
Weighted Loss = - \sum^{N}_{i} {w_i} {L_i} \log({S_i})
\end{equation}
where ${w_i}$ is the ${N}$-vector classes weights. There is a weight for each class in which the non-dominant classes take higher weight than dominant classes which means more penalization. Weights of classes are calculated based on the frequency of each class in the training dataset based on its point sets number as in the following equation:

\begin{equation}
\label{weighted_loss_equations}
Class (i) weights = \log \Big( 1 + \small{\frac { Class(i) Points} {All Classes Points}}  + \epsilon \Big) ^ {-1}
\end{equation}
where $one$ is added in the $\log$ function in order to avoid negative weights, and $\epsilon$ is a small value in order avoid zero weight in case of zero class $(i)$ points.



\subsubsection{Self-Incremental Transfer Learning}
\label{Self_Incremental_Section}
Transfer learning \cite{pan2010survey} is one of the most common techniques in deep learning nowadays. Instead of start learning the model from scratch, neural networks start from a pre-trained network's model rather than using random weights initialization.

Self-Incremental transfer learning is proposed to solve dealing with these datasets through the procedure described in Figure \ref{fig:Incremental_Learning} and as follows: a) modifying the training dataset to have only the non-dominant fine-grained classes, then b) training the model until having a stable saved model, after that c) adding one of the dominant coarse-grained classes, then d) training the same model with loading the pre-trained model saved at in step (b) until having a stable saved model, and finally e) repeating the previous steps until having representative generalized model for the imbalanced dataset.

\begin{figure*}
\centering
\fbox{\includegraphics[width=1\textwidth]{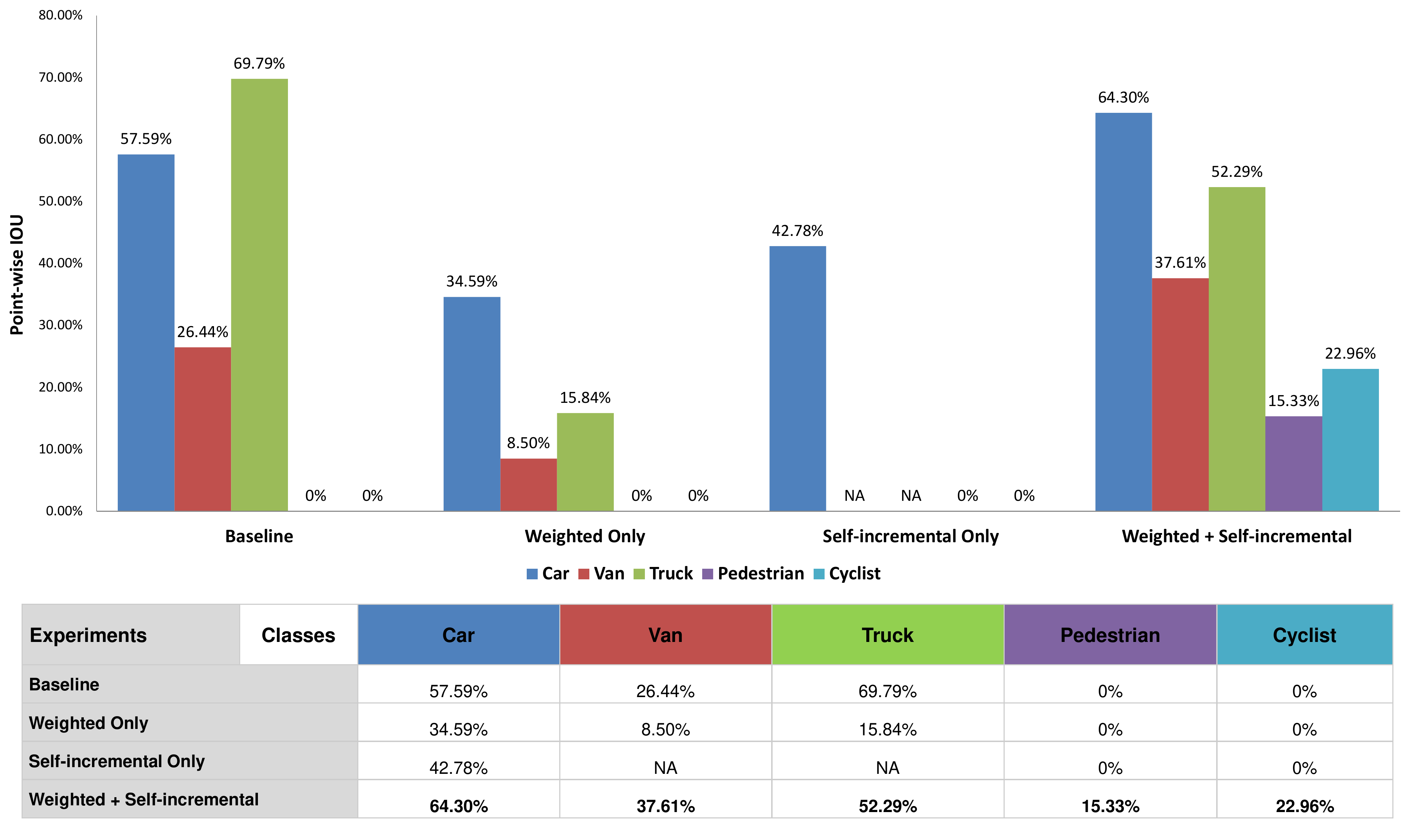}}
\caption{PointNet++ Experiments on KITTI Dataset}
\label{fig:PointNet++_Experiments}
\end{figure*}

\section{Experimental Setup}
\label{Exper_Res}
This section describes the experimental setup, the dataset used, and the main measurement metric for the experiments conducted.


\subsection{Dataset}  
KITTI dataset \cite{Geiger2013IJRR} is most commonly used in autonomous driving research field. We depend on Velodyne Laser Scanner that have 360$^{\circ}$ scanpoints describing the surrounded environment. Unfortunately, KITTI dataset, up till now, does not have a semantic segmentation benchmark. Accordingly, instead of point segmentation, the 3D bounding box is used in order to label the 3D point cloud to have 3D ground truth points. Scanpoints are filtered to have only 180$^{\circ}$ scene in front of the ego-vehicle. The ground reflections add a lot of noise to the point cloud, and it is not affecting the objects of interest, so the points below a certain height are filtered out in the preprocessing phase.

\subsection{Measurement Metric}
Intersection over Union (IoU) per class is the main measurement metric for all experiments. Our proposed IoU is different from the normal Mean IoU used in semantic segmentation for 2D images \cite{barkau1996unet} due to the nature of unsorted 3D point cloud. It is calculated using the correctly classified points, the incorrectly classified points, and the misclassified points. IoU per class is calculated based on the following equation:

\begin{equation}
IoU Class(i) = \frac{TP}{TP + FN + FP}
\end{equation}
where $TP$ is the total number of true positive points which are correctly predicted as $class(i)$, $FN$ is the total number of false negative points which belong to $class(i)$, but they are predicted as another classes, and $FP$ is the total number of false positive points which are predicted as $class(i)$, but they belong to another classes.

\begin{table}
\renewcommand{\arraystretch}{1.3}
\begin{center}
\caption{KITTI Classes Loss Weights}
\label{Weighted_loss_weights}
\centering
\begin{tabular}{|c||c|}
    \hline
    Class  &  Weight\\
    \hline
    No object   & 	1.469		\\
    \hline
    Cars        & 	16.306		\\
    \hline
    Trucks      & 	16.306		\\
    \hline
    Vans	    & 	16.306		\\
    \hline
    Pedestrians & 	48.749		\\
    \hline
    Cyclists	& 	48.604		\\
    \hline
\end{tabular}
\end{center}
\end{table}

\section{Experimental Results}
\label{Exper_Results}
This section describes all conducted experiments comparing between four different methods: baseline, weighted loss only, self-incremental transfer learning only, and our proposed methodology weighted self-incremental transfer learning. We focus on the most important classes in KITTI dataset: car, truck, van, pedestrian and cyclist classes. KITTI dataset is an imbalanced point cloud dataset as it is biased towards: car, van, and truck classes respectively compared with pedestrian, and cyclist classes. According to our statistical analysis, total number of points for pedestrians and cyclists combined together represents approximately  $< 4\%$ of the number of points for car class only.


\subsection{Baseline Experiment}
It is the base of our experiments in which PointNet++ \cite{haopointnet++} is trained on KITTI dataset. However, it fails in classifying KITTI non-dominant fine-grained classes like: pedestrians, and cyclists. This experiment allows us to monitor the main problem of imbalanced point cloud datasets.


\subsection{Weighted Loss only Experiment}
Weighted Loss is considered as the first experiment to solve imbalanced point cloud datasets problems. The cross entropy loss of PointNet++ architecture is replaced by weighted cross entropy loss which is described in section \ref{Weighted_Loss_Function} depending on the weights calculated in table \ref{Weighted_loss_weights}. However, it also fails in classifying KITTI non-dominant fine-grained classes: pedestrians, and cyclists. This experiment shows that the weighted loss function alone does not have the ability to solve imbalanced point cloud datasets problems.



\subsection{Self-Incremental Transfer Learning only Experiment}
Self-Incremental transfer learning is considered as the second experiment to solve imbalanced point cloud datasets problems. We kept using the normal cross entropy loss of PointNet++, and followed the self-incremental transfer learning technique as a learning procedure described \ref{Self_Incremental_Section}. The first training phase runs only on pedestrian, and cyclist classes considering others as no object classes. In the second phase, car class is added to the training dataset depending on the pre-trained model from the first phase. Unfortunately, the model forgets pedestrians and cyclists classes due to the dominance of the car class, so we do not resume the procedure of adding truck and van classes, this is why (N/A) is written in Figure \ref{fig:PointNet++_Experiments}. This experiment also shows that following the self-incremental transfer learning procedure alone does not have the ability to solve imbalanced point cloud datasets problems.

\begin{figure*}
\centering
\fbox{\includegraphics[width=1\textwidth]{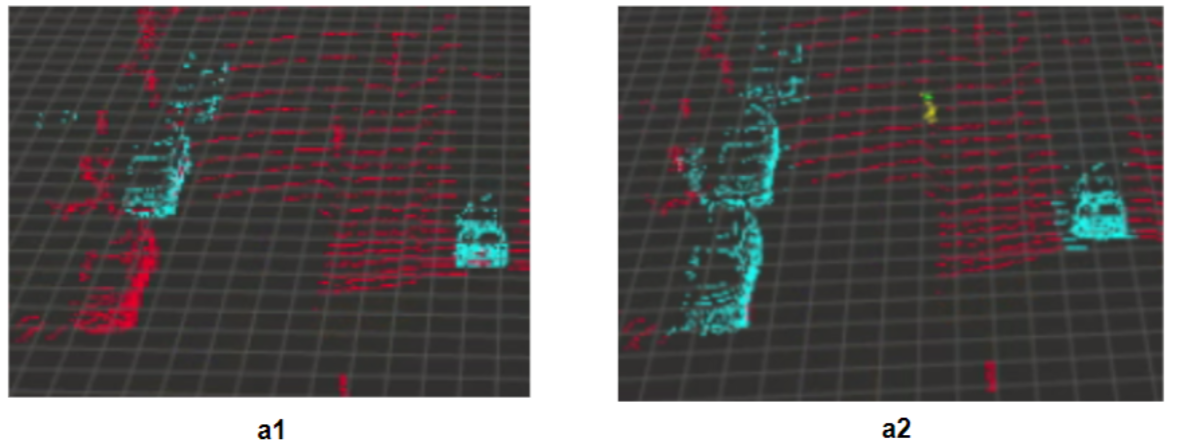}}
\caption{Example of Ground truth labeled scene on the left (a1) vs. 3D point cloud Semantic Segmentation on the right (a2) }
\label{fig:kitti_benchmark}
\end{figure*}

\subsection{Weighted Self-Incremental Transfer Learning Experiment}
Here, our proposed methodology Weighted Self-Incremental transfer learning merges both weighted loss function with the self-incremental transfer learning together.

The first training phase runs only on pedestrian, and cyclist classes considering the others as no object class, and taking the calculated weights in table \ref{Weighted_loss_weights} into account. In the second phase, Car class is added to the training dataset depending on the calculated weights, and continuing learning using pre-trained model from the first phase. Likewise, in the third phase truck and van classes are added to the training dataset depending on the calculated weights, and continuing learning using pre-trained model from the second phase. This experiment shows that our methodology solves imbalanced point cloud datasets problems. Mean IoU measurement metric for the whole experiments are calculated as shown in Figure \ref{fig:PointNet++_Experiments} where \textit{Baseline}, \textit{Weighted Only}, \textit{Self-Incremental Only}, and \textit{Weighted + Self-incremental} are stated in the table below the Figure where (N/A) means that the experiment is not conducted on truck and van classes.

As shown in Figure \ref{fig:PointNet++_Experiments}, the mean IOU for car class is improved from \textit{Baseline} experiment to the \textit{Weighted + Self-incremental} experiment by $6.71\%$, van class is improved by $11.17\%$, while pedestrian and cyclist classes were not recognized or classified in the \textit{Baseline} experiment, but now they are recognized and classified after using our methodology with mean IoU $15.33\%$, and $22.96\%$ respectively. However, truck class is probably decreased by $17.5\%$ due to some confusion with car and van classes. As a result, the overall mean IOU improvement is $39.2\%$. These results illustrate that our methodology not only solves imbalanced point cloud datasets problems, but also enhances the overall accuracy of point-wise classification. 


As shown in Figure \ref{fig:kitti_benchmark}, our proposed methodology provides a 3D semantic segmentation benchmark for KITTI dataset. It shows a semantic segmentation output prediction example vs. the ground truth scene labeling. The proposed methodology ensures generalization for the learned model which is clear from comparing $a1$ with $a2$ scenes. There are missed annotation points in the Ground truth scenes while our model predicts and classifies them correctly.

\section{Conclusions}
\label{conclusion}
Safe Autonomous Driving depends on accurate and robust perception of the surrounding environment. In this work we propose an accurate perception through 3D scene labeling based on the Weighted Self-Incremental Transfer Learning hybrid methodology for solving the imbalanced biased training dataset. Based on the experimental results, it is clear that weighted Loss function alone, and self-incremental transfer learning alone is not enough to solve the imbalanced datasets like KITTI. Our methodology is not only a learning way for the fine-grained classes but also enhances the classification accuracy. It is considered as a generalized methodology that can be used in 3D-semantic segmentation, and different computer vision tasks. It reaches a stable model that does not forget the firstly learned fine-grained classes. Our experimental results introduce a 3D point cloud semantic segmentation benchmark for KITTI dataset.

\addtolength{\textheight}{-12cm}   





\bibliography{Bibtex.bib} 
\bibliographystyle{IEEEtran}

\end{document}